\documentclass[10pt,twocolumn,letterpaper]{article}

\usepackage[pagenumbers]{cvpr} 

\definecolor{cvprblue}{rgb}{0.21,0.49,0.74}
\usepackage[pagebackref,breaklinks,colorlinks,allcolors=cvprblue]{hyperref}
\usepackage{multirow}

\def\paperID{7123} 
\def\confName{CVPR}
\def\confYear{2025}

\title{Schedule On the Fly: Diffusion Time Prediction for \\ Faster and Better Image Generation}

\author{Zilyu Ye$^{1,2,\dagger}$\quad Zhiyang Chen$^{1,4,*}$\quad Tiancheng Li$^{1,2}$\quad Zemin Huang$^{1}$\quad Weijian Luo$^{1,3}$\quad Guo-Jun Qi$^{1,4,*}$\\
    $^1$MAPLE Lab, Westlake University\\
    $^2$South China University of Technology, $^3$Peking University\\
    $^4$Institute of Advanced Technology, Westlake Institute for Advanced Study\\
    \tt\small zilyuye@foxmail.com, \{chenzhiyang, litiancheng, huangzemin\}@westlake.edu.cn\\
\tt\small luoweijian@stu.pku.edu.cn, guojunq@gmail.com
\\
 \url{https://github.com/maple-research-lab/TPDM}
}

\begin{document}
\maketitle
\let\thefootnote\relax\footnotemark\footnotetext{The project was supported by MAPLE Lab at Westlake University.}
\let\thefootnote\relax\footnotetext{$\dagger$ The work was done during an internship at MAPLE lab.}
\let\thefootnote\relax\footnotetext{$^*$ Corresponding author. G.-J. Qi conceived and formulated the idea.}

\begin{abstract}
Diffusion and flow matching models have achieved remarkable success in text-to-image generation. However, these models typically rely on the predetermined denoising schedules for all prompts. The multi-step reverse diffusion process can be regarded as a kind of chain-of-thought for generating high-quality images step by step. Therefore, diffusion models should reason for each instance to adaptively determine the optimal noise schedule, achieving high generation quality with sampling efficiency. In this paper, we introduce the Time Prediction Diffusion Model (TPDM) for this.
TPDM employs a plug-and-play Time Prediction Module (TPM) that predicts the next noise level based on current latent features at each denoising step. 
We train the TPM using reinforcement learning to maximize a reward that encourages high final image quality while penalizing excessive denoising steps.
With such an adaptive scheduler, TPDM not only generates high-quality images that are aligned closely with human preferences but also adjusts diffusion time and the number of denoising steps on the fly, enhancing both performance and efficiency. 
With \textbf{Stable Diffusion 3 Medium} architecture, TPDM achieves an aesthetic score of \textbf{5.44} and a human preference score (HPS) of \textbf{29.59}, while using around 50\% fewer denoising steps to achieve better performance.
\end{abstract}    
\section{Introduction}
\label{sec:intro}

In recent years, deep generative models, including diffusion models \citep{song2021scorebasedgenerativemodelingstochastic,ho2020denoisingdiffusionprobabilisticmodels,sohldickstein2015deepunsupervisedlearningusing} have achieved extraordinary performance across a variety of tasks, including image synthesis \cite{saharia2022photorealistic,ramesh2022hierarchical,ramesh2021zero,karras2020analyzing,karras2022edm}, video generation \cite{ho2022video,blattmann2023stablevideodiffusionscaling,yang2024cogvideoxtexttovideodiffusionmodels}, and others \cite{kong2021diffwave,oord2016wavenet,shen2023difftalkcraftingdiffusionmodels,luo2024entropy}. As a multi-step denoising framework, diffusion models progressively refine random noise into coherent data through iterative sampling, which underlies their impressive capabilities in generating high-quality, diverse outputs.

Inference with a diffusion model involves selecting a noise scheduler, e.g. how the noise level changes step by step when denoising from Gaussian noise. 
This naturally forms a kind of chain-of-thought \cite{wei2022chain}, as it determines how the model gradually generates a real image.
Many works reveal that different noise schedules could greatly affect the model performance \cite{karras2022edm}.
The leading flow-matching models, Stable Diffusion 3~\cite{esserScalingRectifiedFlow2024} and FLUX~\cite{AnnouncingBlackForest2024}, provide recommended noise schedulers according to the targets' resolution. Sabour \cite{sabourAlignYourSteps2024a} and Xia \cite{xiaMoreAccurateDiffusion2023} explore ways to fine-tune the schedule for a model, improving either its efficiency or overall performance. In addition, some one-step generators \citep{huang2024flow,Luo2023DiffInstructAU,luo2024diffinstruct,luo2024one} also achieve impressive performance.
Despite the excellent performance they achieve, most of these works hold the assumption that there exists a universally applicable schedule that is optimal for all prompts and images, which is doubtful.
\begin{figure*}
    \centering
    \includegraphics[width=1.0\linewidth]{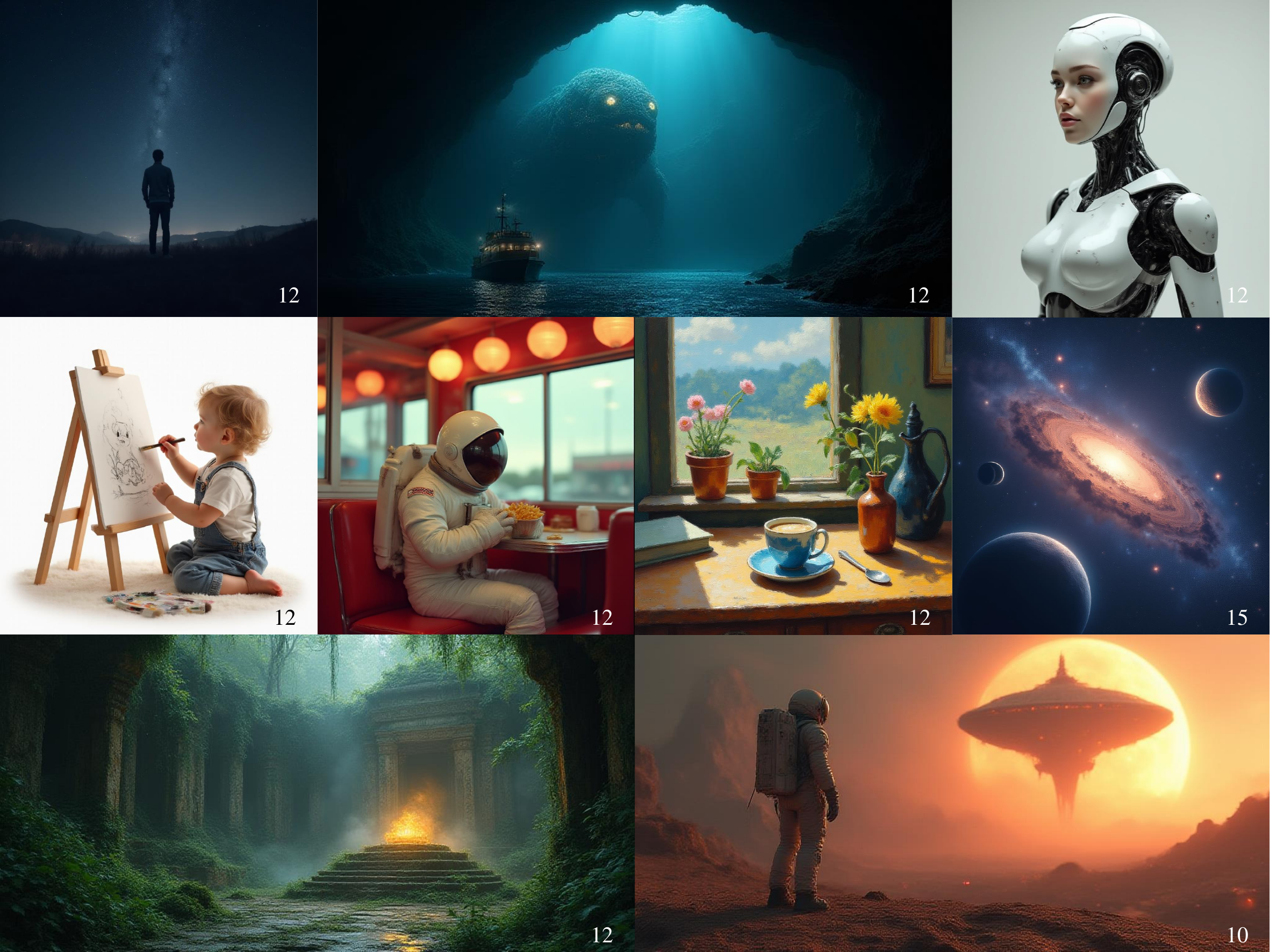}
    \caption{Samples generated by TPDM-FLUX.1-dev showcase stunning visual effects while adaptively adjusting inference steps based on the target output. The number in the lower right corner of each image indicates the inference steps used.}
    \label{fig:first_pic}
    \vspace{-3mm}
\end{figure*}

Let us see some examples in Fig.~\ref{fig:schedule}. The rightmost images contain more complex visual structures generated from longer prompts. These examples make it necessary for the diffusion model to engage a longer chain of sampling steps to generate richer contents. In contrast, the images on the left have simpler structures (e.g., a single object and uniform background) that can be generated with a shorter chain of fewer steps. 
Moreover, as discussed later, the noise level set by diffusion time $t$ for each step matters a lot in properly determining how a sample should be denoised at various steps to eventually generate a high quality image.
Thus, we ask: \emph{Is it possible to adaptively determine both the number of denoising steps and the noise level of each step for each chain of reverse diffusion steps leading to its generated image?} 

In this paper, we propose Time Prediction Diffusion Models (TPDMs) that can adaptively adjust the noise level at each step as well as the total number of steps during inference.
This is achieved by a plug-and-play Time Prediction Module (TPM) that can predict the next diffusion time conditioned on the latent features at the current step.
By regarding the reverse diffusion process as a chain of multiple decisions on the diffusion time at each step, we implement reinforcement learning to maximize the reward computed by a reward model aligned with human preference~\cite{xu2023imagerewardlearningevaluatinghuman}.
The reward does not only reflect the quality of the generated image, but is also discounted by the number of steps used in the generation process.
Thus, TPDM is directly optimized to generate high-quality images with reduced denoising steps.
Moreover, TPM can be easily integrated into any diffusion models with marginal additional computation, allowing them to dynamically adjust the noise schedule for better performance and efficiency.

We implement TPDM on several state-of-the-art models, including Stable Diffusion 3 and Flux. With adaptive noise schedulers, TPDM can generate images with \textbf{$50\%$} fewer steps on average, while keeping the quality on par or slightly better (\textbf{0.322} CLIP-T, \textbf{5.445} Aesthetic Score, \textbf{22.33} Pick Score, \textbf{29.59} HPS v2.1) than Stable Diffusion 3.
We visualize some generated images in Fig.~\ref{fig:first_pic} as well as the number of steps used.
These results demonstrate that TPDM has the potential to either pursue high-quality image generation or improve model efficiency.

Our contributions are summarized below:
\begin{itemize} 
    \item We introduce the Time Prediction Diffusion Model (TPDM). It can predict next diffusion time at each denoising step, determining the optimal noise schedule for each inference instance. 
    \item We train TPDM with reinforcement learning and maximize a reward with regard to both the quality of the generated image and the number of steps, directly optimizing the final performance and efficiency. 
    \item TPDM demonstrates improved performance across multiple evaluation benchmarks, achieving better results with fewer inference steps.
\end{itemize}
\section{Related Works}
\label{sec:rela_works}
\subsection{Diffusion Models}
Diffusion Probability Models (DPMs) \citep{ho2020denoisingdiffusionprobabilisticmodels,sohldickstein2015deepunsupervisedlearningusing} recover the original data from pure Gaussian noise by learning the data distribution across varying noise levels. With their strong adaptability to complex data distributions, diffusion models have achieved remarkable performance in diverse fields such as images \citep{nichol2021improved,oord2016wavenet,poole2022dreamfusion,hoogeboom2022equivariant,kim2021guidedtts},
videos \citep{ho2022video,blattmann2023stablevideodiffusionscaling,yang2024cogvideoxtexttovideodiffusionmodels,liu2024dynamicscaler}, and others \citep{zhang2023enhancing,wang2024integrating,deng2024variational,ye2024openstorylargescaledatasetbenchmark,liu2024r3cd}, significantly advancing the capabilities of Artificial Intelligence Generated Content.
\subsection{Noise Schedule}
In order to generate an image, the model must determine the diffusion time for each step. This can be achieved using either discrete-time schedulers \cite{ho2020denoisingdiffusionprobabilisticmodels,song2022denoisingdiffusionimplicitmodels} or continuous-time schedulers \cite{liu2022flowstraightfastlearning,lipmanFlowMatchingGenerative2023} depending on the model. Typically, the diffusion time reflects the noise level at each step, and most existing approaches rely on pre-determined schedulers. Currently, the leading flow-matching models, Stable Diffusion 3\cite{esserScalingRectifiedFlow2024} and FLUX\cite{AnnouncingBlackForest2024}, provide recommended schedulers that adjust noise levels only based on the target resolution.

Some methods optimize the scheduler to speed up sampling or improve image quality. \citet{xiaMoreAccurateDiffusion2023} predict a new diffusion time for each step to find a more accurate integration direction, \citet{sabourAlignYourSteps2024a} use stochastic calculus to find the optimal sampling plan for different schedulers and different models. \citet{10.1145/3580305.3599412} leverage reinforcement learning to automatically search for an optimal sampling scheduler. Also, adjusting the scheduler is also effective in other areas such as molecular generation~\cite{jung2024conditionalsynthesis3dmolecules}. Some one-step generators\citep{luo2024diffinstruct,luo2024one,huang2024flow} with diffusion distillation \citep{luo2023comprehensive} also achieve impressive performance.

All of the above noise schedulers not only have high thresholds that require users to adjust some hyper-parameters, but also use the same denoising schedule for all prompts and images. On the contrary, TPDM can adaptively adjust the noise schedule during inference and select the optimal sampling plan with suitable sampling steps for each generation, enhancing image quality and model efficiency. We will introduce the detailed definitions and practical algorithms of TPDM in Sec.~\ref{sec:methods}.
\subsection{Reinforcement Learning and Learning from Human Feedback}
Reinforcement Learning from Human Feedback (RLHF) has recently gained significant attention in the field of large language models (LLMs)~\cite{schulmanProximalPolicyOptimization2017, ahmadianBackBasicsRevisiting2024, NEURIPS2020_1f89885d} and is gradually expanding into other domains. Advances in diffusion models have increasingly incorporated reward models to enhance alignment with human preferences~\cite{liangRichHumanFeedback2024, xu2023imagerewardlearningevaluatinghuman, wuHumanPreferenceScore2023}.
By treating latents as actions, diffusion models can be optimized with policy gradient (DDPO \cite{blackTrainingDiffusionModels2023}, DPOK\cite{fan2023dpokreinforcementlearningfinetuning}), actor-critic framework (DACER \cite{wang2024diffusionactorcriticentropyregulator}), direct preference optimization (Diffusion-DPO \cite{wallace2023diffusionmodelalignmentusing}) and other reinforcement algorithms.
Some recent works have also studied the RLHF for few-step generative models \citep{luo2024diffpp,luo2024diffstar}.
Since these approaches rely on the Gaussian reverse processes of the diffusion SDE sampler, however, they are challenging to apply in flow-matching models. On the contrary, in this paper we treat the diffusion time as the action instead of latents. Our goal is to adjust the schedule to achieve better quality with fewer steps, offering a more general and flexible solution.
\section{The Proposed Approach}
\label{sec:methods}
In this section, we first briefly review the fundamental principles of diffusion models, followed by an introduction to the Time Prediction Diffusion Model (TPDM). Finally, we detail the TPDM's training algorithm.

\begin{figure*}
    \centering
    \hfill
    \includegraphics[width=\textwidth]{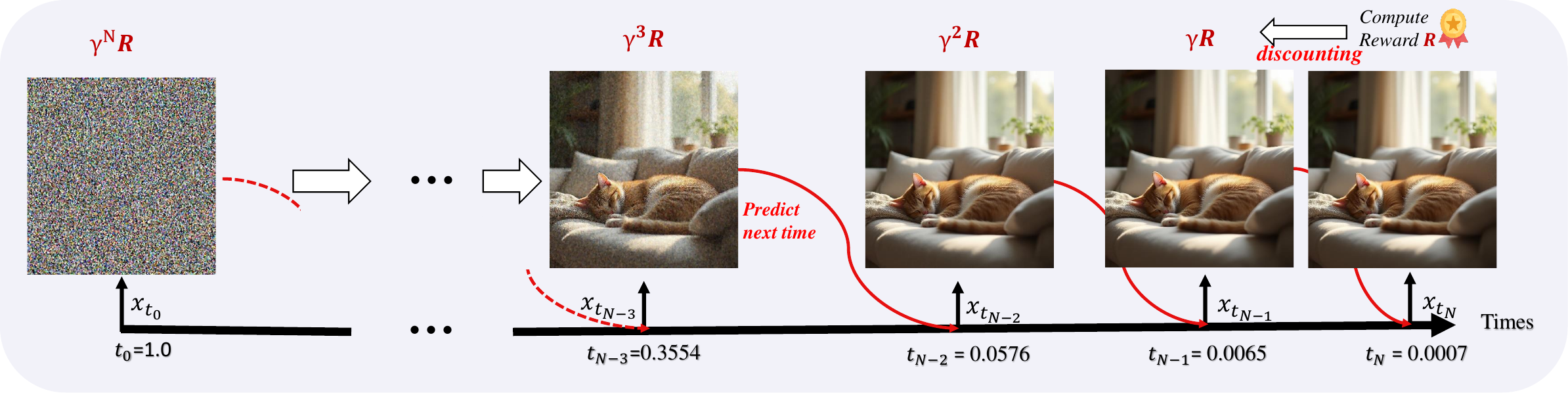}
    \caption{The Inference Process of TPDMs: The horizontal axis represents diffusion time, ranging from 1 to 0. The image starts from random noise $ x_{t_0} $ and is progressively denoised until a clean image $ x_{t_N} $. Meanwhile, the reward is calculated for the final image and discounted by $ \gamma $ to influence previous steps.}
    \label{fig:denosing}
    \vspace{-3mm}
\end{figure*}

\begin{figure}
    \centering
    \includegraphics[width=\linewidth]{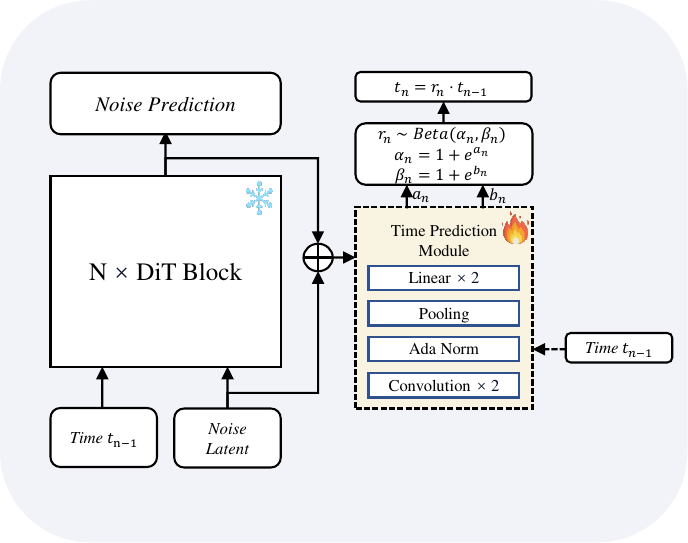}
    \caption{The architecture of TPDM involves a frozen Diffusion Models, and a plug-and-play Time Prediction Module.}
    \label{fig:architecture}
    \vspace{-3mm}
\end{figure}

\subsection{Preliminary}
Diffusion models learn to generate images with a reverse process that gradually removes noise from a sample. The leading paradigm is flow matching~\cite{lipmanFlowMatchingGenerative2023,esserScalingRectifiedFlow2024}. Thus, we introduce how flow-matching models work, and the detailed structure inside current state-of-the-art models here.

We consider a generative model that establishes a mapping between samples $x_1$ drawn from a noise distribution $p_1$ and samples $x_0$ from a data distribution $p_0$. The Flow Matching objective aims to directly regress a vector field $v_t$, which generates a probability flow, enabling the transition from $p_1$ to $p_0$.

\begin{equation}
    \mathcal{L}_{FM}(\theta) = \mathbb{E}_{t, p_t(x)}  \left\| v_{\theta}(x_t,t) - u(x_t,t) \right\|^2 
\end{equation}

The flow-matching model with parameters $\theta$ aims to predict the noise prediction function $ v_{\theta}(x_t,t)$, which approximates the true velocity field $ u(x_t,t)$ that guides the diffusion process from the noise distribution to the clean distribution of generated samples.
Thus, we can get the diffusion ODE:
\begin{equation}
\frac{dx_t}{dt} = v_\theta(x_t, t)
\end{equation}

During inference, suppose we generate an image with $N$ steps, each step has a time $t_n$ corresponding to its noise level.
Then, the $n$-th generation step can be formulated as:
\begin{equation}
x_{t_{n}}=x_{t_{n-1}}+(t_{n}-t_{n-1}) \cdot v_\theta(x_{t_{n-1}}, t_{n-1})
\label{eq:denoise}
\end{equation}
In other words, before the clean image $x_{t_N}$ is finally generated, the model forms a chain of $\{x_{t_n}\}_{n=1}^{N-1}$ as intermediate results. Usually, in a typical flow matching algorithm, there is a pre-determined schedule for each $t_n$, and the proposed TPDM will dynamically decide it for individual samples.

Currently, many state-of-the-art diffusion models are built upon the DiT architecture~\cite{AnnouncingBlackForest2024, chenPixArt$a$FastTraining2023, esserScalingRectifiedFlow2024}. They employ multiple layers of transformers to condition the network on both the diffusion time and the text prompts during denoising steps. In this paper, we will build our TPDM on DiT for the sake of a fair comparison.

\subsection{Time Prediction Diffusion Model (TPDM)}
\label{subsec:TPDM}
To enable the model to adjust the noise schedule on the fly, the TPDM predicts the next diffusion time at each denoising step, as shown in Fig.~\ref{fig:denosing}.
This can be done by adding a lightweight Time Prediction Module (TPM) to the diffusion model as shown in Fig.~\ref{fig:architecture}. This module concatenates the latent features before and after the DiT blocks as inputs, so that both the original noisy inputs and the denoised results at this step are taken into consideration. Then, after several convolution layers, TPM pools the latent features into a single feature vector for prediction. Moreover, we also use an adaptive normalization layer~\cite{perez2017filmvisualreasoninggeneral} in TPM so that the model is aware of the current diffusion time.

Diffusion time sets the noise level, which should monotonically decrease throughout the denoising process. To prevent backward progression, at each step, TPM predicts decay rate $r$ instead, which quantifies how much diffusion time $t$ decreases between adjacent steps. Since the reinforcement learning-based training algorithm in Sec.~\ref{sec:method_train} requires a probability of the predicted time, the final stage of TPM employs two linear layers to process the pooled features and predict the parameters of distribution of $r$ rather than predicting a deterministic value.

Suppose that we are at the $n$-th denoising step. The distribution of the decay rate $r_{n}$ is modeled as a Beta distribution over $(0, 1)$ with two parameters $\alpha_n$ and $\beta_n$ predicted by TPDM. We note that ensuring $\alpha_n>1$ and $\beta_n>1$ results in a unimodal distribution, which is desired as it avoids TPDM from sampling a potentially vague decay rate from a multi-mode beta distribution. To enforce this constraint, we reparameterize the model such that TPDM predicts two real-valued parameters, $a_n$ and $b_n$, from which $\alpha_n$ and $\beta_n$ are determined using Eq.~\ref{eq:alpha_beta}. Consequently, the decay rate $r_{n}$ and the next diffusion time $t_{n}$ can be sampled as in Eq.~\ref{eq:r} and Eq.~\ref{eq:t_2}.

\begin{gather}
\alpha_n = 1 + e^{a_n}, \qquad \beta_n  = 1 + e^{b_n}, \label{eq:alpha_beta} \\
r_n \sim \text{Beta}(\alpha_n, \beta_n)\label{eq:r} \\
t_{n} = r_n \cdot t_{n-1}\label{eq:t_2}
\end{gather}

During training, we freeze the original diffusion model and only update the newly introduced TPM. Thus, the model learns to predict the next diffusion time while preserving the original capacity for image generation.

\begin{figure*}[htbp]
    \centering
    \begin{minipage}[b]{0.95\linewidth}
        \centering
        \includegraphics[width=\linewidth]{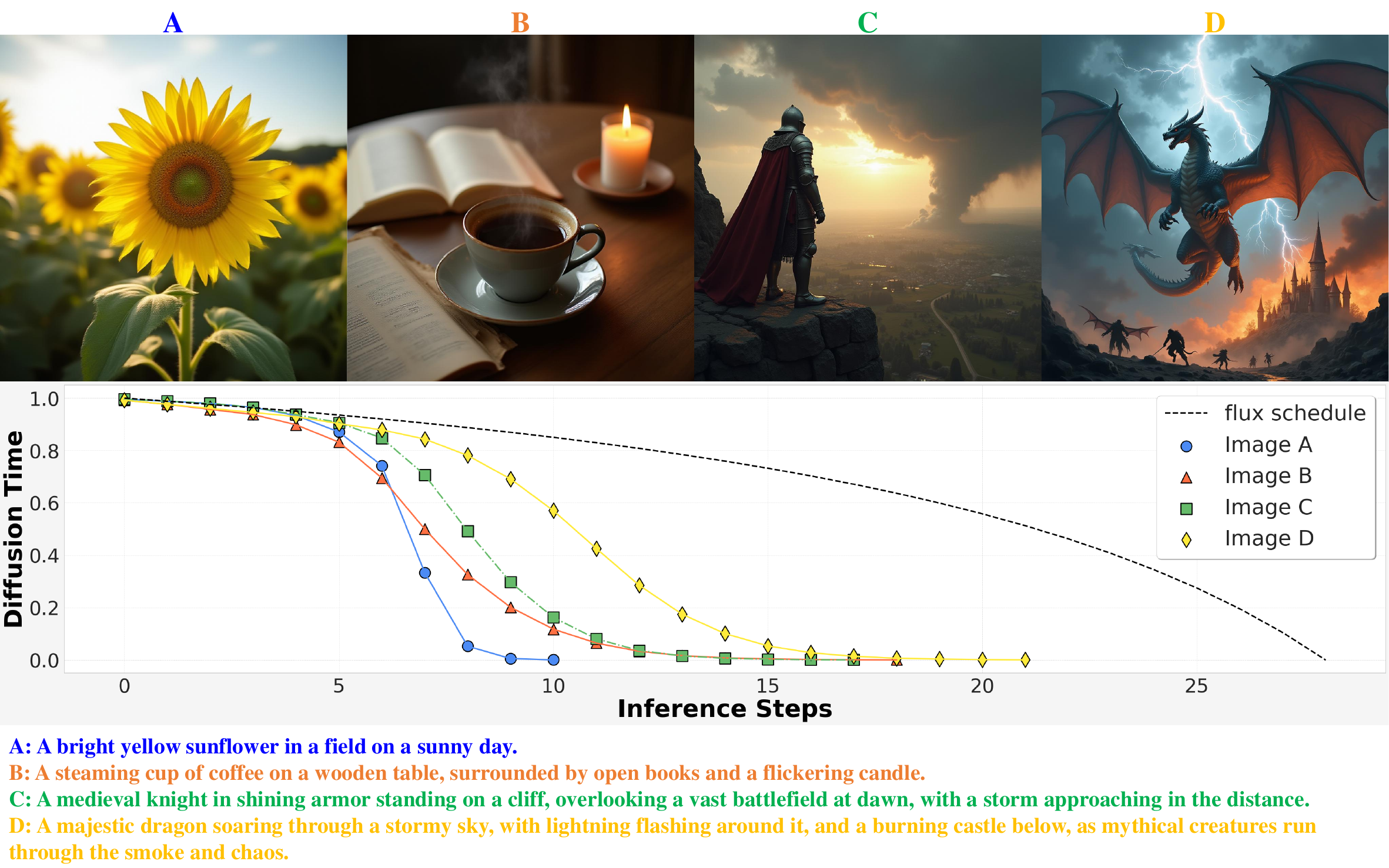}

        \label{fig:schedule1}
    \end{minipage}
    \begin{minipage}[b]{0.95\linewidth}
        \centering
        \includegraphics[width=\linewidth]{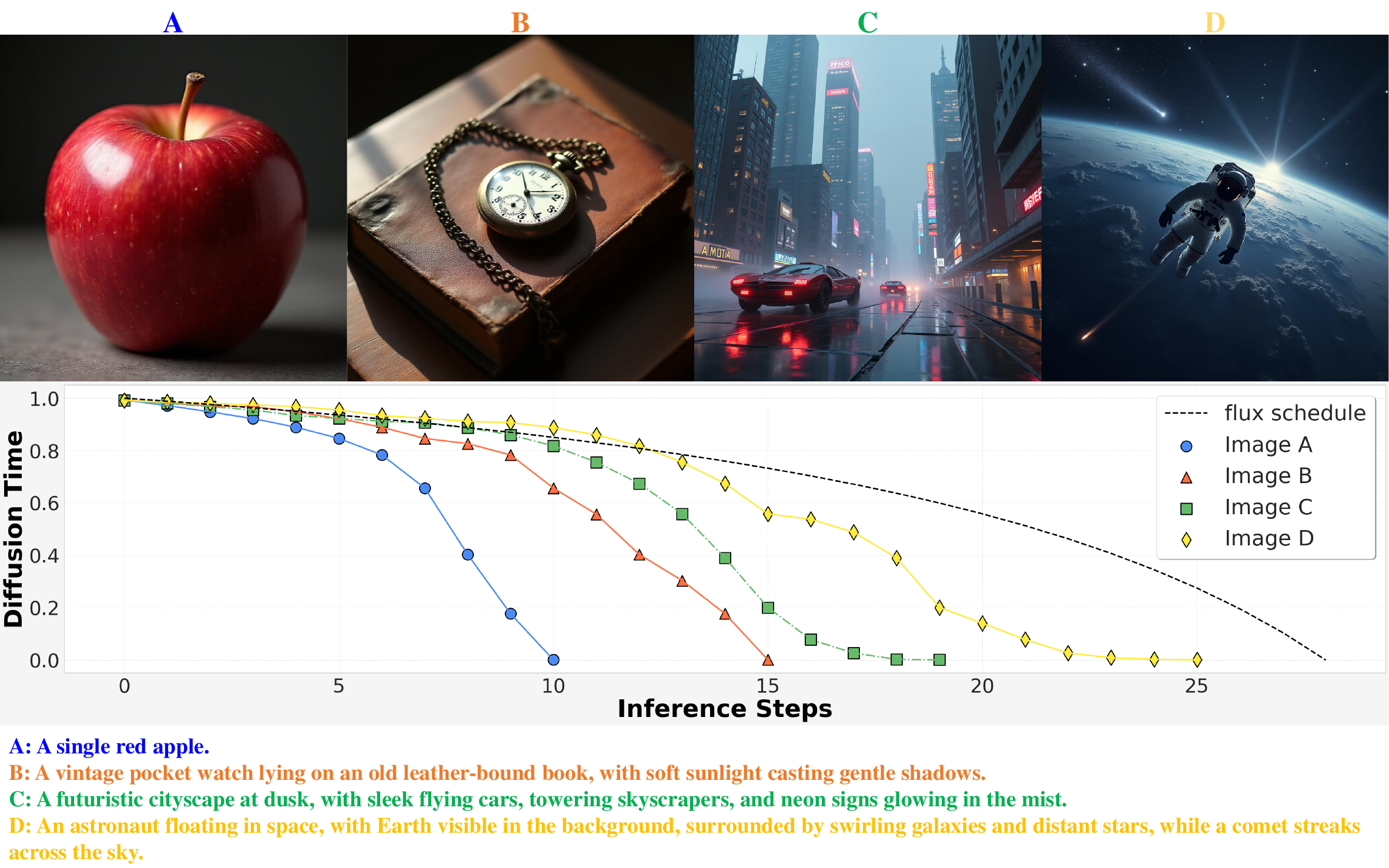}
        \label{fig:schedule2}
    \end{minipage}
    \caption{From left to right, the images generated by TPDM-FLUX1.0-dev progress from simple to complex. Our Time Prediction Module adaptively adjusts the generation schedule to suit the complexity of each generation target.}
    \label{fig:schedule}
    \vspace{-3mm}
\end{figure*}

\begin{figure*}
    \centering
    \includegraphics[width=1.0\linewidth]{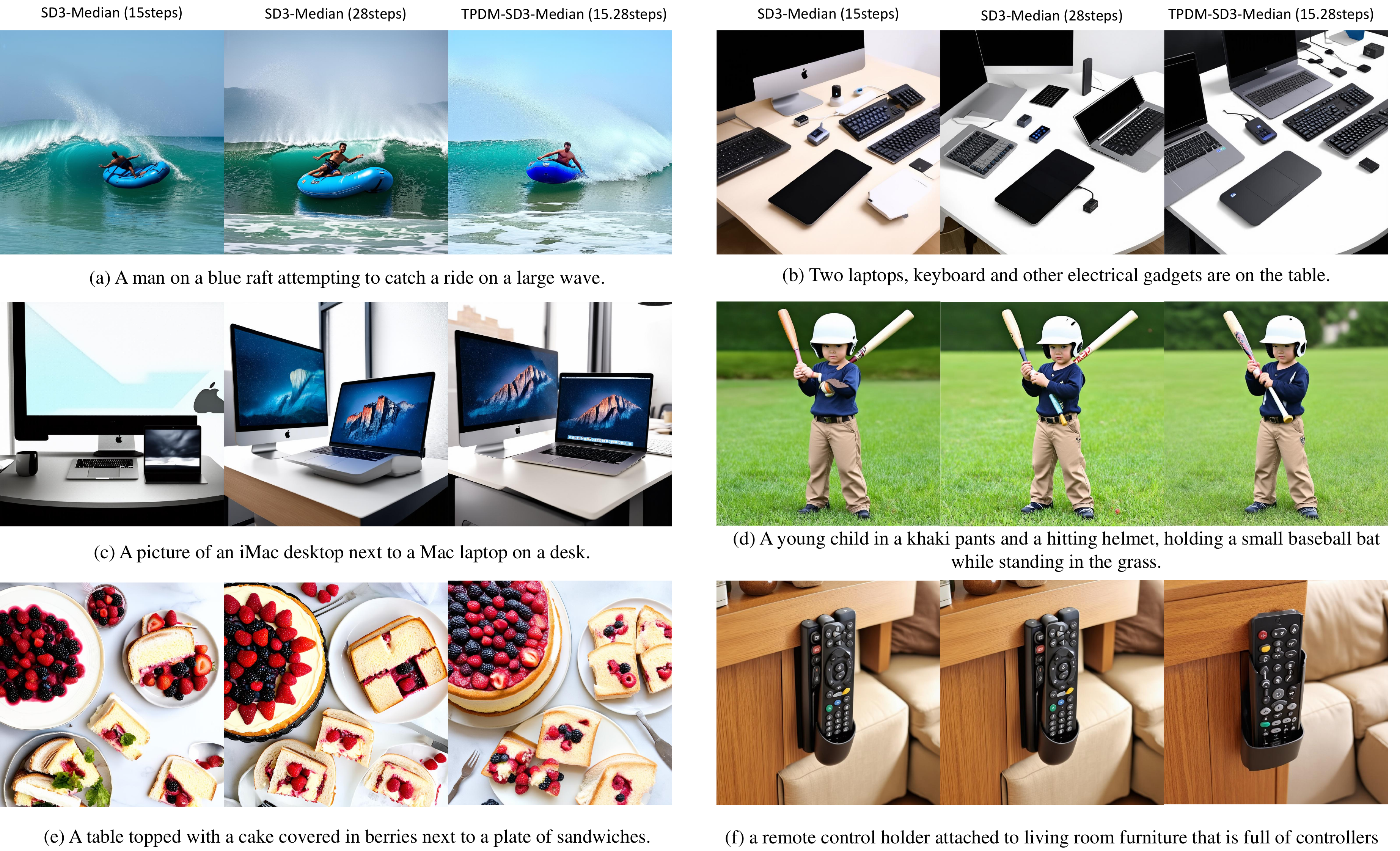}
    \caption{Our TPDM-SD3-Medium, when compared to the SD3-Medium with the recommended and equivalent number of steps, demonstrates superior detail processing ability and generation accuracy.}
    \label{fig:compare}
    \vspace{-3mm}
\end{figure*}

\subsection{Training Algorithms}
\label{sec:method_train}
To train TPM, we need to roll out at least two denoising steps: predicting the next diffusion time with the first step, and denoising with it at the next step. A naive method provides a noised image as input to the first step and trains the model with the reconstruction loss calculated at the second step. The gradients would backpropagate through the predicted $t_n$ to update TPM in the first step. 
However, we found that the trained model tends to complete the denoising process in very few steps during inference, leading to poor image quality.
We hypothesize that by supervising the loss calculated after two steps, the model learns to generate a fully denoised image after two steps, and stop intermediately to minimize the loss function. However, what matters is the final image generated after the whole diffusion reverse process, which is ignored in this method.

After summarizing the above failure, we optimize TPM to maximize the image quality generated after the whole denoising process to achieve precise time prediction. The quality is measured by an image reward model~\cite{xu2023imagerewardlearningevaluatinghuman}.
Considering the computational graph of whole inference is too deep for gradients to backpropagate, so we train the model with Proximal Policy Optimization (PPO) \cite{schulmanProximalPolicyOptimization2017}, whose loss function is formulated as
\begin{equation}
\mathcal{L}(\theta) = -\left[\frac{\pi_{\theta}(y|s)}{\pi_{\text{old}}(y|s)} \hat{A}(s, y) - \lambda \text{KL}[\pi_{{\text{ref}}}(\cdot|s), \pi_{\theta}(\cdot|s)] \right]
\label{eq:ppo}
\end{equation}
Under our task, $s=(c, \epsilon)$ denotes initial states including the input prompt $c$ and gaussian noise $\epsilon \sim \mathcal N(0,1)$; $\pi_\theta$ denotes the policy network, e.g. our TPDM model. $\pi_\text{old}$ denotes the old policy for sampling trajectories; $\pi_\text{ref}$ denotes a reference policy for regularization; $y$ denotes the action our policy takes, i.e., the scheduled time; and $\hat{A}(s, y)$ denotes the advantage the action $y$ has given the state $s$.
We will specify them in the following.

\paragraph{Treat the whole schedule as an action}
Usually, when the model makes a sequence of predictions, PPO regards every single prediction as an action and optimizes with them as a batch. Recently, RLOO \cite{ahmadianBackBasicsRevisiting2024} claims that when the reward only appears at the end of the sequence and the state transition is fully deterministic given a sequence element, we can regard the whole sequence as an action with no harm to the performance. Thus for simplicity, the \emph{entire generation}, including all predicted time in the schedule, is considered as a single action to optimize.

Thus, when computing expectation in Eq.~\ref{eq:ppo}, we consider the whole trajectory as a training sample in optimization. However, TPDM only outputs the distribution of each single time prediction, which we denote by $\pi^{(n)}_\theta$ for the $n$-th step. By the chain rule, the probability of the entire generation can be calculated as the product of each prediction:
\begin{equation}
   \pi_\theta(t_1, ..., t_N | s)  = \prod_{n=1}^{N} \pi^{(n)}_\theta(t_n|s, t_1, ..., t_{n-1}),
\label{eq:policy}
\end{equation}
where $N$ denotes the total number of the generation steps. In our model, each factor $\pi^{(n)}_\theta$ is computed based on the TPM's output beta distribution at the corresponding step.

\paragraph{Image reward discounted by the number of steps}
When generating samples for policy gradient training, we obtain trajectories by generating images from Gaussian noise with the current TPDM policy. Since there exists no ground truth for these generated images, we choose ImageReward~\cite{xu2023imagerewardlearningevaluatinghuman} to assign a reward score based solely on the final image in the last denoising step.
We aim to enhance final image quality without endlessly increasing denoising steps, so we apply a discount factor $\gamma < 1$ (a hyper-parameter) to discount reward to intermediate diffusion steps and calculate the average over the total number of denoising steps $N$. The final reward for this trajectory is shown in Eq.~\ref{eq:reward} below
\begin{equation}
R(s,y) = \frac{1}{N}\sum_{n=1}^{N} \gamma^{N-n} \text{IR}(y, s)
\label{eq:reward}
\end{equation}
where $\text{IR}(y, s)$ denotes the reward of the image generated according to the initial state $s$ and the predicted action $y$.
We refer readers to RLOO \cite{ahmadianBackBasicsRevisiting2024} on how to compute the advantage $\hat{A}(s, y)$ directly from a batch of $R(s,y)$ without a value model for optimization.

This reward function would encourage TPDM to generate images with higher quality and more efficiently. 
By adjusting $\gamma$, we can control how fast the reward calculated at the final step decays when propagating to previous steps. If a smaller $\gamma$ is set, the reward will decay faster as the number of steps increases.
Thus, the model learns to allocate fewer steps in general. We will elaborate more in Sec.~\ref{sec:main_result}.
\section{Experiments}
\label{sec:experiments}

\subsection{Implementation details}
\paragraph{Dataset}
We collect text prompts to train our model. These prompts are either from COCO 2017\cite{coco} train set or by captioning images in Laion-Art~\cite{Laion-Art} and COYO-700M~\cite{kakaobrain2022coyo-700m} datasets with Florence-2~\cite{xiao2023florence} and Llava-Next~\cite{liu2024llavanext}. We will elaborate on this in Appendix~\ref{sec:data_details}.
\paragraph{Training Configurations}\vspace{-5mm}
We utilize the AdamW optimizer with beta coefficients of \((0.9, 0.99)\), a constant learning rate of \(1 \times 10^{-5}\), and a maximum gradient norm of 1.0.
Our TPM module typically requires only 200 training steps. In each step, we sample a batch size of 256 trajectories, and update the model parameter 4 times with them.
\paragraph{Evaluation Metrics}
\vspace{-5mm}
We use FID, CLIP-T, Aesthetic v2, and Pick Score as evaluation metrics by testing with 5,000 prompts from COCO 2017 validation set. For HPS v2.1, we use the 3,200 prompts provided by the benchmark \cite{wuHumanPreferenceScore2023}.

\subsection{Main Results}\label{sec:main_result}
\paragraph{Adaptive Schedule for Different Images}
In Fig.~\ref{fig:schedule}, we present images generated with different prompts and their corresponding schedules predicted by TPDM. When prompting TPDM with shorter and simpler prompts, it requires fewer objects and details to appear in the generated image. Thus the diffusion time decreases faster and reaches $0$ in relatively fewer steps. On the contrary, when more complex prompts are provided, the model needs to reason with more intermediate steps, so that it can generate more delicate details at early and middle stages. Therefore, the diffusion time decreases more slowly. In this case, TPDM requires more denoising steps in the generation process. In the Appendix~\ref{sec:analysis_scedule}, we provide more examples of predicted schedules for analysis.

\begin{table*}[t]
\centering
\small
    \begin{tabular}{c|c c c c c c}
    \hline
    \multirow{2}{*}{Models} & \multirow{2}{*}{Inference Steps} & \multirow{2}{*}{FID} & \multirow{2}{*}{CLIP-T} & \multirow{2}{*}{Aesthetic v2} & \multirow{2}{*}{Pick Score} & \multirow{2}{*}{HPSv2.1} \\
    & & & & & & \\
    \hline
    SD3-Medium~\cite{esserScalingRectifiedFlow2024} & 28 & \underline{25.00} & \textbf{0.322} & \underline{5.433} & 22.12 & \underline{29.12}\\
    SD3-Medium & 15 & \textbf{24.72} & 0.321 & 5.426 & \underline{22.30} & {28.52} \\
    TPDM-SD3-Medium & 15.28 & 25.26 &\textbf{0.322} & \textbf{5.445} & \textbf{22.33} & \textbf{29.59}\\
    \hline
    SD3.5-Large~\cite{StableDiffusion35} & 28 & \textbf{23.29} & 0.318 & 5.487 & \textbf{22.81} & \textbf{30.85}\\
    SD3.5-Large & 15 & \underline{23.35} & \textbf{0.323} & 5.475 & 22.62 & 30.02 \\
    TPDM-SD3.5-Large & 15.22 & 24.48 & \underline{0.322} & \textbf{5.525} & \textbf{22.81} & \underline{30.64} \\
    \hline
    FLUX.1-dev~\cite{AnnouncingBlackForest2024} & 28 & \underline{29.09} & 0.308 & \underline{5.622} & \textbf{23.03} & \textbf{31.94} \\
    FLUX.1-dev & 15 & 29.10 & 0.306 & 5.613 & 22.86 & \underline{31.14} \\
    TPDM-FLUX.1-dev & 13.57 & \textbf{28.98} & \textbf{0.314} & \textbf{5.685} & \underline{22.94} & 30.77 \\
    \hline
    \end{tabular}
\caption{Evaluation of models across various benchmarks, The best result is highlighted in bold, and the second best result is underlined}
\label{tab:table1}
\vspace{-3mm}
\end{table*}

\begin{figure}
    \centering
    \includegraphics[width=0.8\linewidth]{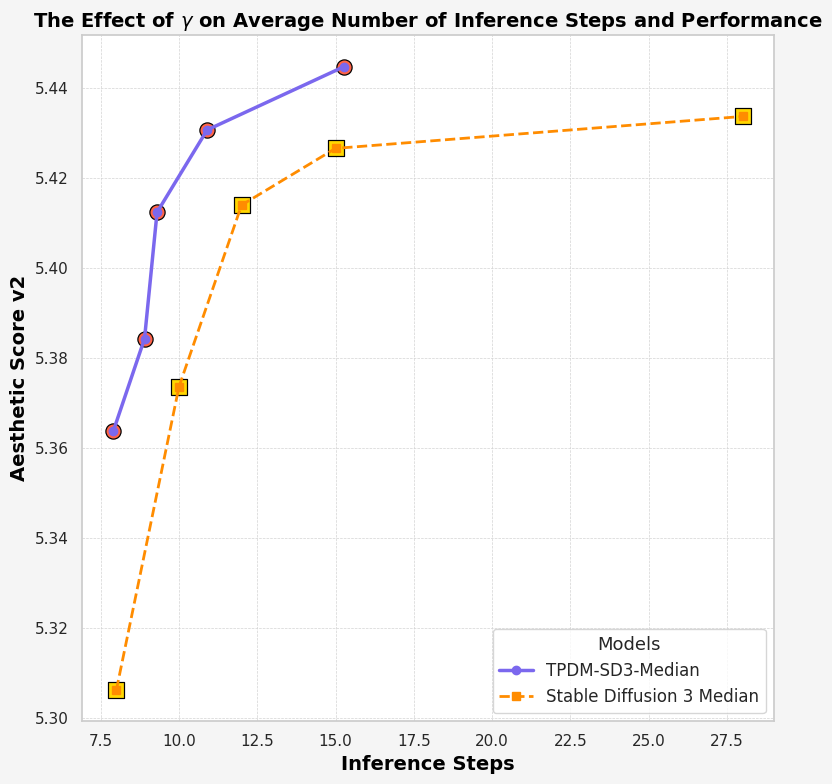}
    \caption{The effect of $\gamma$ on the average number of inference steps.}
    \label{fig:gamma}
    \vspace{-5mm}
\end{figure}
\paragraph{Adjusting $\gamma$ for Different Number of Steps}\vspace{-5mm}
$\gamma$ in Eq.~\ref{eq:reward} controls how fast the reward decays when propagating from the final step to intermediate steps, as shown in Fig.~\ref{fig:denosing}, thereby affecting the average number of denosing steps of our converged TPDM.

As shown in Fig.~\ref{fig:gamma}, as we decrease $\gamma$ from 0.97 to 0.85, TPDM tends to decrease the diffusion time more rapidly, leading to fewer denosing steps, from 15.0 to 7.5. Additionally, when compared to the baseline (yellow line), TPDM (purple line) consistently achieves a significantly higher aesthetic score with the same number of inference steps.

\paragraph{Visual Comparison}\vspace{-5mm}
Fig.\ref{fig:compare} provides some specific images generated by both TPDM and SD3-Medium. In Fig.~\ref{fig:compare}(c), the image generated by TPDM have a more realistic laptop keyboard than SD3-Medium with 15 and 28 steps. As for Fig.\ref{fig:compare}(d), TPDM removes the extra baseball bat, leading to a more natural looking. Generally, TPDM is able to generate more accurate and realistic images.

\subsection{Quantitative Results}
We apply TPM on several state-of-the-art diffusion models, including SD3-Medium, SD3.5-Large, and Flux 1.0 dev, and demonstrate how TPM could enhance their performance. Other than commly used metrics like FID, CLIP-T and human preference scores (Aesthetic Score v2, and HPSv2.1), we provide a user study as well.

\paragraph{Quantitative Metrics}\vspace{-5mm}
We compared TPDM and the above-mentioned models in Tab.~\ref{tab:table1}.
While maintaining competitive performance, TPDM can generate images in about half of their recommended steps on average.

Among all the metrics, those representing human preference improve the most. For example, by generating with just 15.28 steps on average, TPDM-SD3-Medium gets a 29.59 HPS score, +1.07 higher than Stable Diffusion 3 with similar steps, also +0.47 higher than the original 28-step results. This improvement may be attributed to the reward model, which is well aligned with human preferences. By optimizing image generation based on this alignment, TPDM produces more aesthetically pleasing results.
Furthermore, since other scheduler optimization methods mostly conduct experiments on classical diffusion models such as Stable Diffusion 1.5~\cite{rombach2022highresolutionimagesynthesislatent}, we conduct an apple-to-apple comparison with them in Appendix~\ref{sec:tpdm_ddpm} as well.

\paragraph{User Study}\vspace{-5mm}
To better reflect human preference towards these models, we conduct a user study by inviting volunteers to compare images generated from different models.

Specifically, for each prompt, we provide two images generated by SD3-Medium with 15 and 28 steps separately, and an image generated by TPDM-SD3-Medium. We generate 50 groups of images for comparison in total, and invite 15 volunteers to select the ones they prefer. The result is shown in Tab.~\ref{tab:table_human}, indicating that our model can generate images that better align human preferences.

\begin{table}
\centering
\small
    \begin{tabular}{c|c c c c}
    \hline
    Models & Inference Steps & Win Rate \\
    \hline
    SD3-Medium & 28 & 26.58\% \\
    SD3-Medium & 15 & 16.40\% \\
    TPDM-SD3-Medium & 15.28 & 47.25\% \\
    \hline
    \end{tabular}
\caption{User study based on Stable Diffusion 3 architecture}
\label{tab:table_human}
\vspace{-3mm}
\end{table}

\paragraph{Ablation on the Architecture}\vspace{-5mm}
Tab.~\ref{tab:input_hidden} indicates that taking features from both the first and last layer as the inputs for TPM performs better than only taking features from either the beginning or last two layers.

\begin{table}
\setlength{\tabcolsep}{1.8pt}
\small
    \centering
    \begin{tabular}{c|c c c c c}
        \hline
        TPM inputs & Steps & FID & CLIP-T & PickScore& Aes v2\\
        \hline
        First 2 Layers& 16.20 & 24.81 & 0.321 & 22.18 & 5.400 \\
        Last 2 Layers& 19.30 & \textbf{23.19} & 0.322 & 22.16 & 5.356 \\
        First $\&$ Last Layers& \textbf{15.28} & 25.26 & \textbf{0.322} & \textbf{22.33} & \textbf{5.445} \\
        \hline
    \end{tabular}
    \caption{Ablation Study of inputs hidden states}
    \label{tab:input_hidden}
    \vspace{-3mm}
\end{table}
\section{Conclusions and Limitations}
In this paper, we view denoising steps as a kind of chain-of-thought in image generation, and introduce the Time Prediction Diffusion Model (TPDM) by adaptively predicting the next diffusion time to denoise images in the reverse diffusion process. This adjusts the noise schedule per sample, optimizing image generation for each prompt. By aligning the final outputs of the denoising process with human preferences, TPDM can reduce the number of inference steps by almost 50\% while still keeping image quality.

\noindent \textbf{Limitations.} First, in this paper, we only design a simple architecture for TPM. A more delicate module may yield better performance. Second, we freeze the parameters of the original diffusion model. Updating them iteratively with the training of TPM could lead to improved results, which is left for further exploration.

\noindent \textbf{Acknowledgement.} This work was supported by National Natural Science Foundation of China under Grant No. 92467104, and Zhejiang Leading Innovative and Entrepreneur Team Introduction Program (2024R01007).

{
    \small
    \bibliographystyle{ieeenat_fullname}
    \bibliography{main,mybib}
}

\clearpage

\clearpage
\appendix
\setcounter{page}{1}
\maketitlesupplementary

\section{Dataset Details}
\label{sec:data_details}
In Section 4, we provide a brief overview of our training dataset. Below, we present a more detailed description of the dataset.
\subsection{Dataset Source}
We curated training prompts from three high-quality datasets, COCO \cite{coco}, Laion-art \cite{Laion-Art} and COYO-11M \cite{coyo-hd-11m-llavanext}.

\begin{figure}[h]
    \centering
    \includegraphics[width=0.8\linewidth]{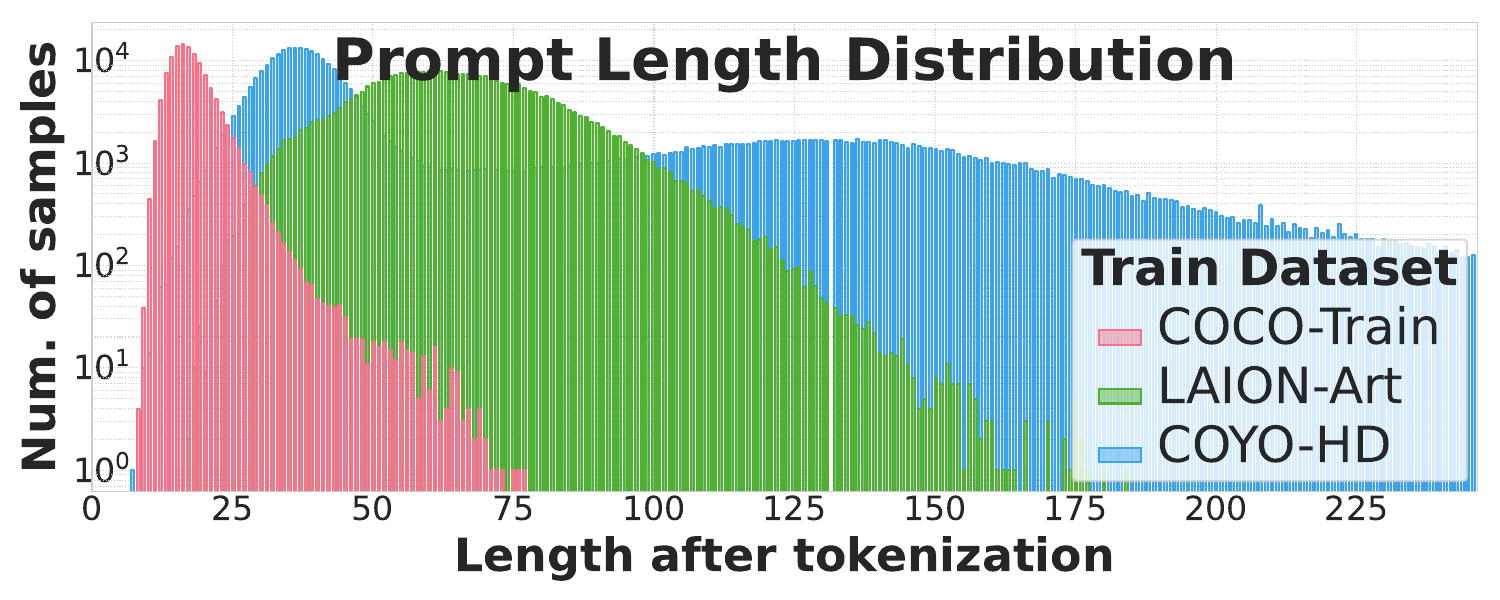} 
    \caption{Prompt length distribution.}
    \label{fig:promptdist}
\end{figure}

For COCO, the original captions in the training split is used.
For Laion-art, we caption the images with Florence-2 \cite{xiao2023florence} to obtain the text prompts. For COYO-11M, we retained its original captioning by Llava-next \cite{liu2024llavanext}. Fig.~\ref{fig:promptdist} illustrates the distribution of prompt length after tokenization.

To ensure diversity during training, we design a filtering pipeline to select a diverse subset of these prompts.

\subsection{Filtering Pipeline}

\begin{figure}[ht]
    \centering
    \includegraphics[width=0.5\linewidth]{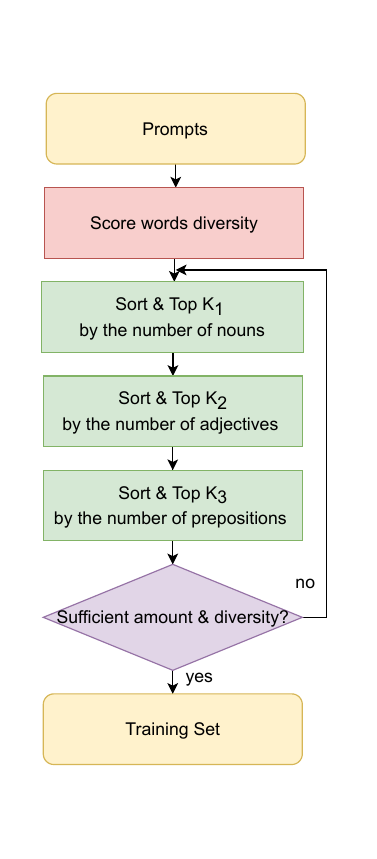}
    \caption{The filtering pipeline used to construct the training set.}
    \label{fig:pipe}
\end{figure}

\begin{figure*}[ht]
    \centering
    \includegraphics[width=1.0\linewidth]{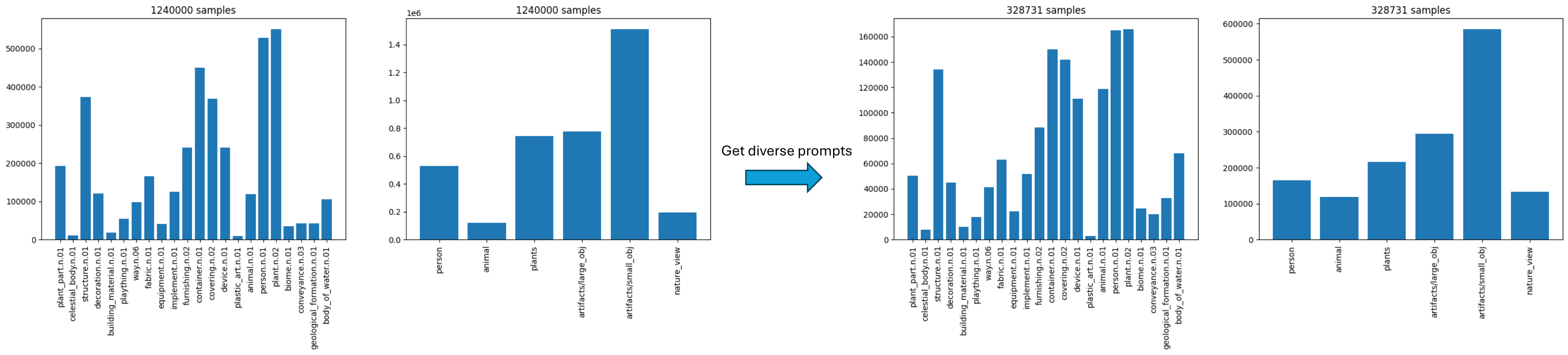}
    \caption{The statistics of prompts from the re-captioned high-resolution Laion-Art dataset, before and after filtering, highlight an improvement in diversity. Our dataset demonstrates greater variety compared to the original.}
    \label{fig:diverse}
\end{figure*}

Our filtering pipeline, depicted in Figure~\ref{fig:pipe}, aims to select diverse and high-quality prompts by analyzing key linguistic features, such as nouns, prepositions, and adjectives. To assess diversity, we utilize \textit{WordNet} \cite{miller1995wordnet} to count the number of valid nouns, adjectives, and prepositions in each prompt.

Prompts are ranked based on these counts. Those exhibiting the highest scores are included in the training set. The selection process prioritizes \textbf{noun-diversity}, ensuring a balanced representation across categories such as \textit{Person}, \textit{Animal}, \textit{Plant}, \textit{Artifacts (Large/Small Objects)}, and \textit{Natural Views}. Next, prompts with high \textbf{preposition-diversity} are selected, emphasizing those that contain spatial and relational terms (e.g., \texttt{near}, \texttt{on}). Finally, prompts are evaluated for \textbf{adjective-diversity}, with particular focus on adjectives describing \textit{color} and \textit{shape}, to enhance descriptive richness.

The process is iterative, selecting the top prompts in order of their noun, adjective, and preposition diversity scores until the desired data quantity (51,200 prompts, as specified in Section 4.1) and diversity are achieved. As shown in Figure~\ref{fig:diverse}, the diversity of the training set has significantly enhanced after filtering.

\begin{table*}[]
    \centering
    \begin{tabular}{l|c c c c c c}
        \hline
        Method & Inference Steps & FID & CLIP-T & Aesthetic v2 & Pick Score & HPSv2.1\\
        \hline
        DPM-Solver++~\cite{lu2023dpmsolverfastsolverguided} & 10 & 21.08 & \textbf{0.314} & \textbf{5.219} & 21.30 & 24.00\\
        DPM-Solver++ \& AYS~\cite{sabourAlignYourSteps2024a}  & 10 & 18.81 & 0.313 & 5.156 & 21.31 & \textbf{24.40} \\
        DPM-Solver++ \& GITS~\cite{chen2024trajectoryregularityodebaseddiffusion} & 10 & 18.32 & 0.313 & 5.108 & 21.19 & 23.72\\
        DPM-Solver++ \& TPM & 9.89 & \textbf{18.31}& 0.313 & 5.118 & \textbf{21.32} & 24.29\\
        \hline
    \end{tabular}
    \caption{Experiments on Stable Diffusion v1.5 \cite{rombach2022highresolutionimagesynthesislatent}, and $9.89$ is the average number of steps TPM used. The best results are highlighted in bold.}
    \label{tab:compare}
\end{table*}

\section{Applicaion on Denoising Diffusion Probabilistic Models (DDPMs)}
\label{sec:tpdm_ddpm}
Before the widespread use of flow-matching models, diffusion models dominated image generation, such as DDPM~\cite{ho2020denoisingdiffusionprobabilisticmodels} and LDM~\cite{rombach2022highresolutionimagesynthesislatent}. They demonstrate several differences from flow models during inference. First, they directly predict the noise added to the data at each step, rather than predicting the velocity vector field. Second, both the diffusion time $t$ and the noise level are typically discrete, unlike continuous time in flow-matching models.

In this section, we integrate TPM into Stable Diffusion v1.5 \cite{rombach2022highresolutionimagesynthesislatent}, demonstrating that TPM also works with DDPMs. At diffusion time $t_{n-1}$, to determine the next diffusion time $t_{n}$, TPM predicts two real-valued parameters, $a_{n}$ and $b_{n}$. These parameters define $\alpha_{n}$ and $\beta_{n}$, which shape the Beta distribution of the diffusion time decay rate $r_{n}$, similar to the flow-matching model version of TPDM introduced in~\ref{subsec:TPDM}. Then $t_{n}$ is obtained using Eq.~\ref{eq:t_2} and quantized to the nearest discrete diffusion time to obtain the corresponding noise level.


We compare the performance of TPDM with other scheduler optimization methods applied to diffusion models. The evaluation metrics are calculated using 5,000 prompts from the COCO 2017 validation set, in line with the evaluation protocol as in Table~\ref{tab:table1}. Notably, our findings reveal that TPM can be combined with higher-order solvers, such as DPM-Solver++ \cite{lu2023dpmsolverfastsolverguided}, resulting in a significant improvement of -2.79 in FID, surpassing the performance of other methods as reported in Tab.~\ref{tab:compare}.

\section{Analysis of the Predicted Schedule}
\label{sec:analysis_scedule}
\begin{figure*}
    \centering
    \includegraphics[width=0.9\linewidth]{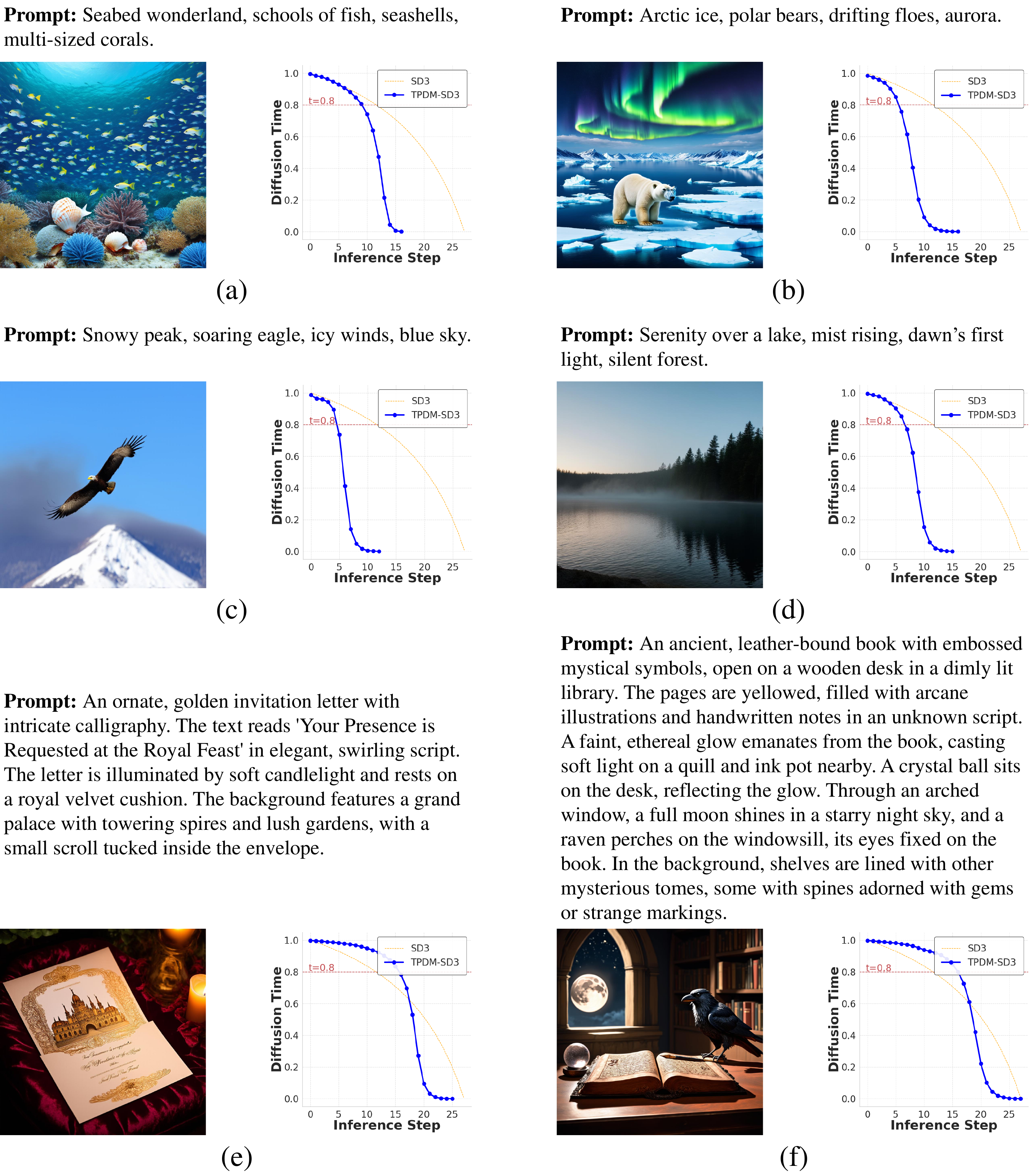}
    \caption{Prompts, images and predicted schedules.}
    \label{fig:more_examples}
\end{figure*}

Fig.~\ref{fig:more_examples} provides more examples to show the noise schedule predicted by TPDM.

We observe that TPDM tends to allocate more inference steps to higher noise levels to generate complex details and layouts. Take Fig.~\ref{fig:more_examples}(a) for an example where multiple objects of various sizes are generated with a complex visual layout -- TPDM allocates 10 out of 17 steps to more noisy diffusion time with $t>0.8$.
This way, TPDM more efficiently spends its inference steps on denoising noisier samples at the earlier stage so that it can add more complex visual layouts and details as early as possible to eventually generate high quality results. It avoids the problem in the benchmark diffusion model that may waste many evenly allocated steps to denoise cleaner images at the later stage. The steep curve in Fig.~\ref{fig:more_examples}(c) indicates that, for simple generation, TPDM reduces the diffusion time much faster to almost zero within only $13$ steps, instead of  evenly allocating $28$ steps in the benchmark model.
Fig.~\ref{fig:more_examples}(e) and \ref{fig:more_examples}(f) also visualize the results for extremely long prompts. They exhibit a similar trend.

\begin{figure}[t]
    \centering
    \includegraphics[width=0.6\linewidth,trim=0 0 0 0.5cm,clip]{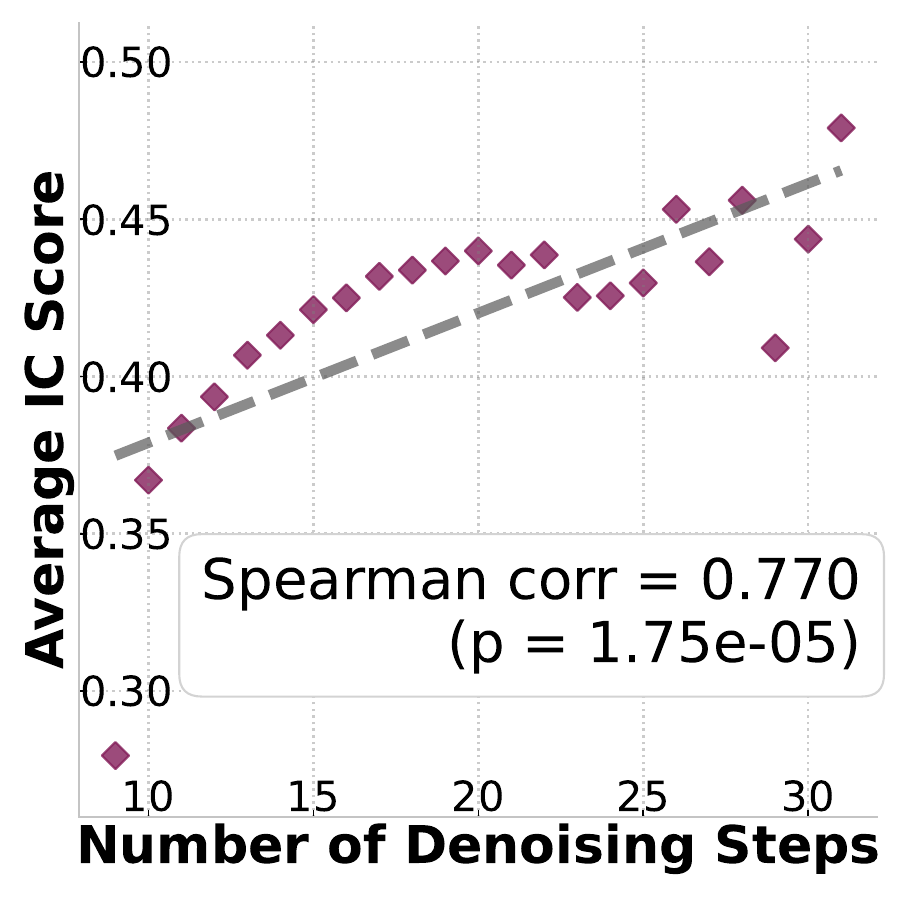}
    \caption{Correlation between the number of steps and image complexity (measured by ICNet \cite{ic9600}).}
    \label{fig:correlation}
\end{figure}

Moreover, Fig.~\ref{fig:schedule} shows that simple prompts lead to faster decrease in diffusion time with fewer steps, while complex ones lead to more steps.
We use ICNet \cite{ic9600}
to score the image complexity, and calculate the average score of images generated with different number of steps in evaluation set. As shown in Fig.~\ref{fig:correlation}, it has a high correlation efficient (0.770), which supports our assumption that complex images are dynamically generated with more steps by TPDM.

\section{Prompts to Generate Images in Fig.~\ref{fig:first_pic}}
\begin{enumerate}
    \item 8k uhd A man looks up at the starry sky, lonely and ethereal, Minimalism, Chaotic composition Op Art.
    \item A deep-sea exploration vessel descending into the pitch-black ocean, its powerful lights illuminating the glowing, alien creatures that inhabit the abyss. A massive, ancient sea creature with bioluminescent patterns drifts into view, its eyes glowing as it watches the explorers from the shadows of an underwater cave.
    \item Half human, half robot, repaired human.
    \item A baby painter trying to draw very simple picture, white background.
    \item an astronaut sitting in a diner, eating fries, cinematic, analog film.
    \item Van Gogh painting of a teacup on the desk.
    \item A galaxy scene with stars, planets, and nebula clouds.
    \item A hidden, forgotten city deep in a jungle, with crumbling stone temples overgrown by thick vines. In the heart of the city, a mysterious glowing artifact lies on an ancient pedestal, surrounded by an eerie mist. Strange symbols shimmer faintly on the stone walls, waiting to be uncovered.
    \item A lone astronaut stranded on a desolate planet, gazing up at the sky. The planet’s surface is cracked and barren, with glowing, unearthly ruins scattered across the horizon. In the distance, a massive, alien ship slowly descends, casting an eerie shadow over the landscape.
\end{enumerate}

\end{document}